\documentclass[runningheads]{llncs}
\usepackage{hyperref}
\usepackage[T1]{fontenc}
\usepackage{multicol}
\usepackage{comment}
\usepackage{graphicx}
\usepackage{comment}
\usepackage{multirow}
\usepackage{booktabs}
\usepackage{makecell}
\usepackage{colortbl}
\usepackage[table,xcdraw]{xcolor}

\begin{document}
\title{Do Superpixel Segmentation Methods Influence Deforestation Image Classification?}
%
%



\author{Hugo Resende\inst{1}\orcidID{0000-0001-9735-905X} \and
Fabio A. Faria\inst{1,2}\orcidID{0000-0003-2956-6326} \and
Eduardo B. Neto\inst{1}\orcidID{0000-0001-6515-0403} \and
Isabela Borlido\inst{3} \orcidID{0000-0001-7288-2485} \and
Victor Sundermann\inst{3} \orcidID{0009-0000-7567-1149} \and
Silvio Jamil F. Guimar\~aes \orcidID{0000-0001-8522-2056} \and
Álvaro L. Fazenda\inst{1}\orcidID{0000-0002-4052-1113} 
}
\authorrunning{Resende et al.}
%
\institute{
Institute of Science and Technology - Universidade Federal de São Paulo,  Avenida Cesare Mansueto Giulio Lattes, n° 1201 - Eugênio de Mello, 12247-014, São José dos Campos, SP, Brazil. \\ 
\email{\{hresende,alvaro.fazenda\}@unifesp.br} \vspace*{0.3cm}
\and 
AIPS, INESC-ID, Instituto Superior Tecnico, Universidade de Lisboa, Portugal. \\
\email{fabio.faria@tecnico.ulisboa.pt} \vspace*{0.3cm}
\and
Computer Science Department, Pontif\'icia Universidade Cat\'olica de Minas Gerais, Belo Horizonte, MG, Brazil. \\
\email{\{isabela.borlido, vgmsundermann\}@sga.pucminas.br, sjamil@pucminas.br}
}


\maketitle             
\begin{abstract}
Image segmentation is a crucial step in various visual applications, including environmental monitoring through remote sensing. In the context of the ForestEyes project, which combines citizen science and machine learning to detect deforestation in tropical forests, image segments are used for labeling by volunteers and subsequent model training. Traditionally, the Simple Linear Iterative Clustering (SLIC) algorithm is adopted as the segmentation method. However, recent studies have indicated that other superpixel-based methods outperform SLIC in remote sensing image segmentation, and might suggest that they are more suitable for the task of detecting deforested areas. In this sense, this study investigated the impact of the four best segmentation methods, together with SLIC, on the training of classifiers for the target application. Initially, the results showed little variation in performance among segmentation methods, even when selecting the top five classifiers using the PyCaret AutoML library. However, by applying a classifier fusion approach (ensemble of classifiers), noticeable improvements in balanced accuracy were observed, highlighting the importance of both the choice of segmentation method and the combination of machine learning-based models for deforestation detection tasks.

\end{abstract}

\section{Introduction}

Image segmentation plays a relevant role in a wide range of applications, including the analysis and interpretation of visual data, such as medical diagnosis, environmental monitoring, precision agriculture, and the generation of data for training machine learning models \cite{azad2024medical,resende2024sampling,charisis2024deep,peng2024advanced,van2024lumbar}. 
Good segmentation accurately identifies homogeneous regions within an image by grouping pixels with similar characteristics. Well-defined segments enable the extraction of more reliable information, reduce noise and inconsistencies in analysis, and support decision-making in automated processes. On the other hand, poorly executed segmentation can lead to the loss of important details, confusion between objects of interest and background or other classes, and significantly compromise the performance of classification or detection models \cite{zhang2022learning,kuchler2024uncertainty}.

Several segmentation strategies have been developed, including deep learning approaches such as semantic, instance, and panoptic segmentation \cite{guo2018semantic,hafiz2020instance,kirillov2019panoptic}. Although these methods achieve high performance in various domains, they require large amounts of annotated data and significant computational resources \cite{minaee2021image}. As an alternative, superpixel-based methods stand out for their simplicity, efficiency, and independence from supervised training. They group similar pixels while preserving important boundaries and structures in the image, making them useful as a preprocessing step. However, they still face challenges such as sensitivity to texture and lighting, as well as difficulty in determining the optimal number of segments \cite{barcelos2024comprehensive}.

The segments generated by superpixel-based methods can be leveraged in citizen science projects focused on environmental monitoring, such as the ForestEyes project \cite{dallaqua2019foresteyes,fazenda_cacm2024}. This initiative combines citizen science and machine learning efforts to monitor tropical forests, particularly in detecting areas showing signs of deforestation. In the project, campaigns are organized on the Zooniverse platform \cite{simpson2014zooniverse,smith2013introduction}, where non-expert volunteers analyze and label segments previously extracted from satellite images. These labeled segments are then used as training data for machine learning models aiming to automate the detection of deforested areas \cite{resende2025increasing}. 
In this context, the choice of the segmentation algorithm is crucial, as it must produce well-defined regions, clearly separate the classes of interest, and generate regular and visually coherent polygons, which helps human cognition and increases the reliability of the labels provided by volunteers.

Designing, creating, and conducting a citizen science campaign are not trivial tasks, as they require careful planning of activities and continuous engagement of volunteers. Each task, represented by a segment to be evaluated, demands time and attention from participants, which can make the process lengthy and costly, especially in campaigns involving numerous segments. In light of this challenge, alternative labeling sources become valuable, such as the official deforestation maps provided by the Amazon Deforestation Monitoring Project (PRODES) \cite{inpe-prodes}. Although these data do not fully replace human contributions, they can help accelerate the evaluation process and provide a reliable reference for training and validating machine learning models.

In the ForestEyes project campaigns, the Simple Linear Iterative Clustering (SLIC) \cite{achanta2012slic} algorithm has been used as the standard segmentation method. However, recent studies by \cite{borlido2024how} and \cite{resende2025igarsstoapper} have shown, through quantitative analyses using both traditional superpixel evaluation metrics and citizen science-based metrics, that among the 22 main superpixel segmentation algorithms found in the literature, there are alternatives that outperform SLIC for the specific goals of the ForestEyes project, suggesting more deeply investigations regarding citizen science and automatic classification. In this context, the present study aimed to investigate the influence of the four best-performing methods identified in those previous studies, along with SLIC, on the training of machine learning models for deforestation detection tasks.

\section{Background}

\begin{figure*}[h!]
\centering
\includegraphics[width=0.9\textwidth]{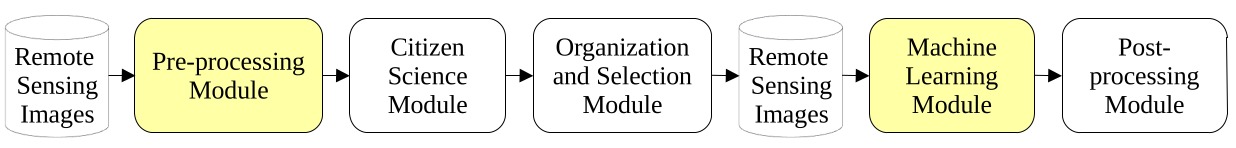}
\caption{The pipeline of the ForestEyes project. Highlighted is the target module of this work.}
\label{fig:FE}
\end{figure*}

In this section, the concepts and main characteristics of the ForestEyes project and the superpixel-based segmentation algorithms, which are the focus of this study, will be presented.

\subsection{ForestEyes Project}

The ForestEyes project is structured around five core modules, as illustrated in Figure \ref{fig:FE}. The \textbf{Pre-processing Module} is responsible for tasks related to the acquisition, preparation, and segmentation of remote sensing data. Satellite imagery is sourced from platforms such as EarthExplorer (for Landsat-8) and the Copernicus Browser (for Sentinel-2). Additionally, deforestation maps from the PRODES project are incorporated, which span from August of one year to July of the next. Data selection requires specifying parameters such as the region of interest and the acquisition period. 
Once collected, the data undergo preparation using tools such a regular GIS. At this stage, non-relevant regions, identified through PRODES, are excluded, enabling the segmentation process to concentrate solely on forested and recently deforested areas. 
The regions of interest are also cropped, and spatial alignment between satellite images and PRODES data is performed to ensure pixel-level correspondence. For segmentation, the project employs a superpixel-based approach. The current method in use is MaskSLIC, though alternative techniques are under evaluation, as discussed in \cite{borlido2024how} and \cite{resende2025igarsstoapper}.

In the \textbf{Citizen Science Module}, classification campaigns are created and deployed on the Zooniverse platform. Each task displays the same image segment using different false-color compositions to help volunteers distinguish between forested and deforested areas more easily. These compositions are selected based on proven methods from the literature and validated by image-processing experts. Metadata is generated to organize the sequential display of tasks and compositions. During campaign setup, the logic of volunteer participation is defined, including parameters such as the minimum number of responses per task, visual layout, instructional content, and usability features. Once launched, campaigns can be monitored in real-time, enabling ongoing support and ensuring the smooth progress of volunteer activity.

In the \textbf{Organization and Selection Module}, responses collected during the citizen science campaigns are analyzed to determine the final classification for each task. The dataset includes metadata such as volunteer identifiers, the tasks completed, and the time taken for each classification. Based on this information, majority voting is applied to derive consensus labels, while response variability is quantified using entropy measures. Response time is also evaluated as a proxy for task complexity.
These metrics are instrumental in identifying inconsistencies, assessing annotation reliability, and informing refinements for future campaigns. Upon completion of this analysis, the module defines the training and testing sets that will be used in the subsequent Machine Learning Module.

The \textbf{Machine Learning Module} consists of a series of steps ranging from the extraction of Haralick texture features from the labeled segments to the application of one or more classification models. With the training and testing sets defined in the previous module, the process begins with feature extraction, which serves as input for the learning phase. Next, a classifier, or a set of classifiers previously used in the ForestEyes project, is trained using these features. Finally, the trained model(s) are evaluated and applied, completing the supervised learning cycle within the context of the project.

Finally, the \textbf{Post-processing Module} is responsible for decision-making activities based on the deforestation detection results obtained in the machine learning module. This module interprets the outcomes and transforms them into actionable information or recommendations. Additionally, future developments are expected to include mechanisms that enable reliable automatic labeling of samples, among other features aimed at enhancing the system’s autonomy and scalability.

\subsection{Superpixel-based Segmentation Algorithms}
\label{sub:superpixels}

The algorithm \textbf{Compactness-Controlled Contour-Relaxed Superpixels (CRS)} is an approach for image pre-segmentation that aims to partition an image into a predefined number of regions (superpixels). The main feature of this method is its ability to simultaneously maximize the internal texture homogeneity within each superpixel and the adherence of contours to the visual boundaries of the image, supported by a Gibbs-Markov random field model. Unlike traditional methods, which limit homogeneity to smooth regions or similar intensity values, CRS remains effective even in highly textured scenes. The algorithm optimizes an energy function, which depends on only a few parameters defined by the chosen statistical model. The deformation of superpixels in response to image content is controlled by a single parameter that regulates the compactness of the segments. Extensive experimental evaluations have shown that CRS outperforms well-established superpixel algorithms in the literature in terms of boundary recall and undersegmentation error while offering competitive or superior runtime performance \cite{CONRAD-2013-CRS}.

\textbf{Eikonal-based Region Growing Clustering (ERGC)} was developed to facilitate the segmentation of anatomical structures in medical images, especially in scenarios that demand speed, such as radiation therapy planning. The technique clusters pixels based on the solution of an Eikonal equation, where each pixel receives a value representing its travel time or geodesic distance from initial seeds. The function that regulates this propagation is computed from the dissimilarity between the pixel features and those of the expanding region, allowing segmentation that is sensitive to the image content. The Fast Marching method is used to accelerate the computation, ensuring efficiency even for large volumes. ERGC also enables control over the compactness of the generated regions, which contributes to the creation of regular and homogeneous segments \cite{BUYSSENS-2014-ERGC}.

\textbf{Efficient Topology Preserving Segmentation (ETPS)} algorithm was developed to provide fast, regular, and topologically coherent segmentation, suitable for applications requiring real-time processing, such as autonomous driving scenarios. It combines a coarse-to-fine energy update strategy, inspired by the SEEDS algorithm, with an efficient optimization approach that significantly reduces the number of boundary updates throughout the process. Unlike previous methods that require multiple iterations to achieve good results, ETPS can reach high-quality energy minima with just a single iteration, while maintaining connectivity between segmented regions. The algorithm also incorporates shape regularization by penalizing boundary length and can integrate additional information, such as stereo data, to enhance segmentation in stereo image pairs. Evaluations on datasets such as BSD and KITTI show that ETPS outperforms popular approaches like SLIC and SEEDS in both segmentation quality and execution speed \cite{YAO-2015-ETPS}.

\textbf{Rooted Spanning Segmentation (RSS)} is a graph-based image segmentation algorithm in which regularly distributed seed pixels serve as roots. Segmentation is performed by extending paths originating from these roots, based on cost functions that consider both the maximum difference and the range of values between pixels. In this way, it balances color similarity and spatial proximity. 
The number and placement of segmented regions are controlled through seed selection, while path-based propagation ensures coherence, contour adherence, and adaptability to image structures. Uniformity and compactness of the resulting regions are modulated by a scaling factor, and connectivity is inherently preserved through the region-growing process.
Among its main features is scalability, as it can handle different types of data in a unified manner. RSS also has promising applications, such as extracting regions based on seeds and cost thresholds, which are useful for tasks like network extraction and global object proposals. Finally, studies also indicate its potential in convolutional neural network-based semantic segmentation, due to its ability to operate on high-dimensional representations extracted by such networks \cite{CHAI-2020-RSS}.

\textbf{Simple Linear Iterative Clustering (SLIC)} is one of the most popular segmentation algorithms due to its simplicity, efficiency, and the quality of its results. It works by grouping pixels based on color similarity and spatial proximity, using a variation of the k-means algorithm. To achieve this, it operates in a combined five-dimensional space, three for color (in the CIELAB space) and two for pixel position, which allows the generation of compact and regular superpixels. One of SLIC's main advantages is the direct control over the number of segmented regions, as well as its ability to produce results that align well with image boundaries. Although the algorithm does not natively guarantee connectivity among pixels within each superpixel, this issue can be addressed with a simple post-processing step. Due to its high computational efficiency, SLIC is widely used as a preprocessing step in various computer vision tasks, such as semantic labeling, tracking, and object proposal generation \cite{achanta2012slic}.

\section{Experiments}

This section presents the details of the experiments conducted in this study. In particular, it describes the construction of the dataset and the results obtained by applying the PyCaret AutoML library and building the classifier ensembles proposed in~\cite{Ferreira2021ensembles}.

\subsection{Dataset Construction}

Like depicted in \cite{resende2024sampling}, \cite{borlido2024how} and 
\cite{resende2025igarsstoapper}
nine remote sensing images from the Landsat-8 satellite (also referred to as study areas) were collected, representing a geographic region of approximately 8,514 hectares, located near the Xingu River Basin in the state of Pará, Brazil. This region is known for its recurrent deforestation hotspots. For each of these images, the corresponding PRODES deforestation map was obtained, matching the acquisition period (year $2022$). After collection, the images underwent a preparation process in a GIS, which included cropping the regions of interest, performing spatial alignment between the Landsat-8 and PRODES data at the pixel level, and creating masks for segmentation. Next, the principal component (PCA) of each image was extracted to serve as the basis for the segmentation step. Based on these preparations, the five superpixel-based segmentation algorithms presented in Subsection \ref{sub:superpixels} were applied, using a fixed parameter of 6,000 superpixels for each method.

\begin{figure}[!ht]
\centering 
\includegraphics[width=0.9\textwidth]{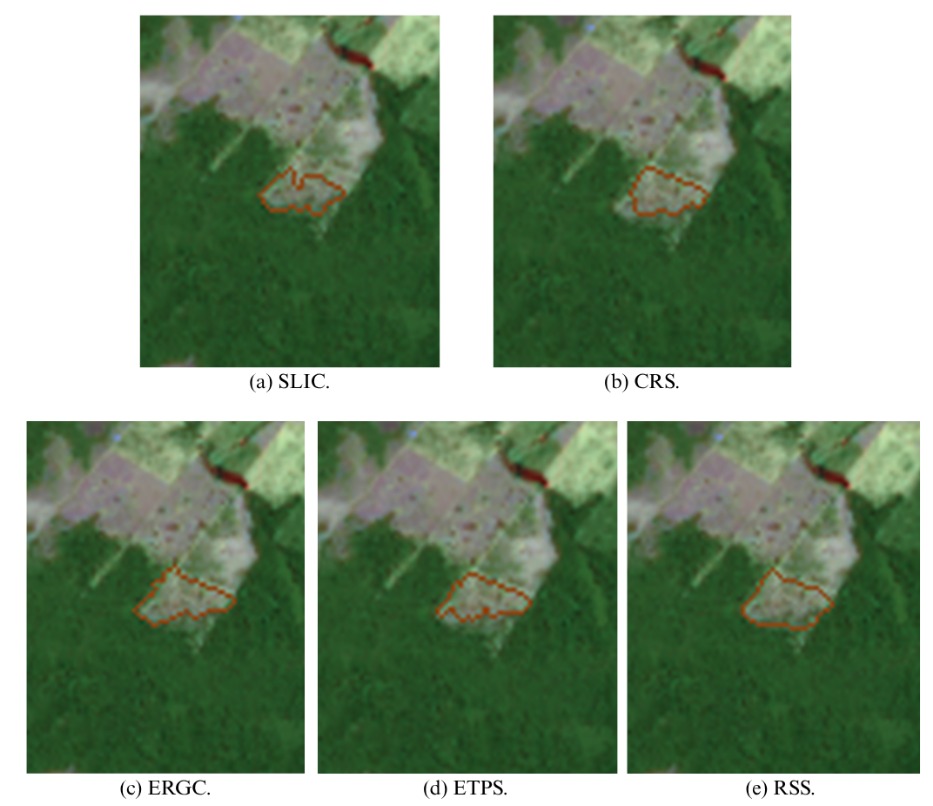}
\caption{Examples of deforestation segment, from the same geographic area, for each one of the segmentation methods. To visualize the segments, the false-color image composed of the $4-6-7$ bands was used to highlight forest and deforestation regions.}
\label{fig:exemploSegmentos}
\end{figure}

Once the segments were generated, a set of useful segments was created for each segmentation method. These segments, as defined in the ForestEyes project, are those containing more than $70$ pixels (given the $30$-meter spatial resolution of Landsat-8) and having over 70\% homogeneity in terms of forest or deforestation pixels. From this set, $45$ segments with perfect homogeneity ($100$\%) were selected for each target class (forest and deforestation), per segmentation method. Additionally, another $45$ segments per class were selected with homogeneity greater than $70$\%, ordered increasingly, to ensure a diverse training set containing both ideal and more heterogeneous examples. As a result, $180$ training segments were selected per segmentation method, totaling $900$ training segments across the five methods. It is important to highlight that, since SLIC is the baseline method used in the ForestEyes project, segments generated by this algorithm were used as reference, and corresponding geographic regions were identified in the outputs of the other segmentation methods (largest intersection of pixels within segments). An example of such equivalent segmented regions is shown in Figure~\ref{fig:exemploSegmentos}.  Finally, the remaining useful segments for each method, those not used in training, were assigned to the respective test sets. Since the dataset was built from nine different study areas and each algorithm produces a different segmentation of the same region, the number of test samples varied slightly among the segmentation methods (CRS$=$19,039, ERGC$=$19,009,  ETPS$=$19,034,  RSS$=$18,999, and SLIC$=$19,044 segments).

After defining the segments for the training and testing sets, 13 Haralick features were extracted for each segment, considering four directions (0º, 45°, 90°, and 135°) in the gray-level co-occurrence matrix for bands 6-4-3-1, following the same experimental protocol adopted in~\cite{neto2024satellite}. For each of the 13 features, the arithmetic mean of the values obtained across the four directions was computed. As a result, each segment was represented by a feature vector containing 13 attributes, which were then used as input for training and evaluating the machine learning models.

\subsection{Results and Discussion}



The \textbf{first stage}, serving as a baseline, of the experiments focused on using the PyCaret AutoML library to identify the best classifiers for each training dataset generated by the different segmentation methods (CRS, ERGC, ETPS, RSS, and SLIC). Initially, the PyCaret automatically evaluates $14$ classification algorithms, including some of the most well-known in the literature, such as Ridge Classifier (RC), Linear Discriminant Analysis	(LDA), Logistic Regression (LR), Gradient Boosting Classifier (GBC), Decision Tree Classifier (DT), Stochastic Gradient Descent Classifier (SGD), Extra Trees Classifier (ET), Light Gradient Boosting Machine (LGBM), Support Vector Machine (SVM), Quadratic Discriminant Analysis (QDA), Random Forest (RF) and k-Nearest Neighbors Classifier (KNN). Each model is assessed using $5$-fold cross-validation, and the results are summarized in a performance table that includes several key metrics such as accuracy, area under the ROC curve (AUC), recall, precision, F1-score, Kappa, MCC, and training time (TT). Based on the aforementioned metrics, PyCaret generates an overall performance ranking, prioritizing classifiers that demonstrate a good balance between accuracy and statistical robustness.

For each superpixel segmentation method, this ranking is used to select the top five machine learning models. These selected classifiers are then submitted to an additional hyperparameter optimization step using PyCaret’s automatic fine-tuning feature, which aims to further enhance model performance. At the end of this process, the best-performing classifiers, with their optimized hyperparameters, are finalized for each dataset associated with each segmentation method. As an example, for the dataset generated by the SLIC segmentation method, the best performing classifier after fine-tuning was Logistic Regression, with the following hyperparameters: C=8.426, class\_weight=`balanced', penalty=`l2', solver=`lbfgs', and max\_iter=1000, among other parameters that remained at their default values. It is noteworthy that the values obtained align with those recommended in the official scikit-learn documentation \cite{scikit-learn}, indicating that the optimization process performed by PyCaret adheres to established best practices and guidelines from the literature.

Once the hyperparameters of the top five classifiers for each segmentation method were tuned, the next step was to test them on the respective test sets associated with each method. Due to the imbalanced classes of the test sets, to evaluate the performance of the machine learning models at this stage, the balanced accuracy metric was adopted. As shown in Table~\ref{tab:top5_classifiers}, the family of linear classifiers, such as Ridge Classifier (RC) and Logistic Regression (LR), stood out as effective classification solutions, regardless of the superpixel method studied in this paper. However, it is important to note that there were no significant differences among the best-performing classifiers, indicating that the evaluated segmentation methods are equally suitable for processing this type of image and for training machine learning models. 

\begin{table}[ht!]
\centering
\caption{Balanced accuracy results (\%) of the top five classifiers for each superpixel method.}
\label{tab:top5_classifiers}
\renewcommand{\arraystretch}{1.1} 
\begin{tabular}{|l|c|c|c|c|c|}
\toprule
\textbf{\makecell{Spx\\Methods}} & \textbf{RC} & \textbf{LDA} & \textbf{LR} & \textbf{GBC} & \textbf{Other} \\
\midrule
CRS   & \textbf{86.00} & 84.91 & 85.78 & 79.87 & DT (73.98) \\ \hline
ERGC  & 84.43 & 84.05 & \textbf{85.50} & 78.07 & SGD (85.13) \\ \hline
ETPS  & \textbf{86.10} & 83.89 & 85.99 & 79.15 & ET (77.89) \\ \hline
RSS   & \textbf{85.10} & 84.27 & 84.62 & --    & \makecell{LGBM (76.96),\\ SGD (84.82)} \\ \hline
SLIC  & 83.92 & 78.88 & \textbf{84.60} & --    & \makecell{DT (77.16),\\ LGBM (78.84)} \\
\bottomrule
\end{tabular}
\end{table}

Given this scenario, the \textbf{second stage} involves selecting the best classifier (top 1) obtained for each superpixel method (CRS, ERGC, ETPS, RSS, and SLIC) and testing it on the test sets of the other segmentation methods. This strategy is referred to as the \textit{cross-method} scenario, and its results can be observed in Table \ref{tab:cross_segmenters_extended}. In this experiment, the classifier names are indicated alongside their corresponding segmentation methods (e.g., RC/CRS indicates that the Ridge Classifier was the best machine learning model for the CRS segmentation method).

\begin{table}[h!]
\centering
\caption{Balanced accuracy (\%) for individual classifiers (top one for each superpixel method) trained on a given superpixel method and tested on others, and performance of ensemble strategies.}
\label{tab:cross_segmenters_extended}
\resizebox{\linewidth}{!}{
\begin{tabular}{|l|c|c|c|c|c|c|c|c|c|c|}
\toprule
\multirow{2}{*}{\textbf{\makecell{Spx}}} & \multicolumn{5}{c|}{\textbf{Individual Classifier/Spx Method}} & \multicolumn{5}{c|}{\textbf{Classifier Ensemble Strategies}} \\ \cline{2-11}

 & \textbf{\makecell{RC/}} & \textbf{\makecell{LR/}} & \textbf{\makecell{RC/}} & \textbf{\makecell{RC/}} & \textbf{\makecell{LR/}} & 
\multicolumn{2}{c|}{\textbf{\makecell{UMDA}}}  &
\multicolumn{2}{c|}{\textbf{\makecell{MV}}} & \textbf{\makecell{Rel. Gain}} \\
\textbf{\makecell{Methods}} &
\textbf{\makecell{CRS}} &
\textbf{\makecell{ERGC}} & 
\textbf{\makecell{ETPS}} &
\textbf{\makecell{RSS}} &
\textbf{\makecell{SLIC}} & 
\textbf{\makecell{Acc.}} &
\textbf{\makecell{$\#$C}} &
\textbf{\makecell{Acc.}} &
\textbf{\makecell{$\#$C}} &
\textbf{\makecell{\tiny{(UMDA$\times$Top1)}}} \\

\midrule
CRS   & 86.00 & 86.60 & \textbf{86.90} & 78.60 & 85.00 & \textbf{88.84} & 12 & 88.50 & 25 & 2.2 \\ \hline
ERGC  & 76.40 & \textbf{85.50} & 79.30 & 84.50 & 84.50 & \textbf{87.38} & 10 & 86.06 & 25 & 2.2 \\ \hline
ETPS  & 83.60 & \textbf{86.60} & 86.10 & 79.40 & 86.30 & \textbf{87.43} & 10 & 86.62 & 25 & 1.0 \\ \hline
RSS   & 77.00 & 82.60 & 80.80 & \textbf{85.10} & 84.20 & \textbf{85.12} & 12 & 83.94 & 25 & 0.0 \\ \hline
SLIC  & 79.10 & \textbf{84.90} & 81.70 & 77.80 & 84.60 & \textbf{86.17} & 14 & 85.07 & 25 & 1.5 \\ 
\specialrule{.20em}{.1em}{.1em}
Avg.  & 80.42 & \textbf{85.24} & 82.96 & 81.08 & 84.92 & \textbf{86.99} & 12 & 86.04 & 25 & -- \\ \hline
Std. Dev. & 4.21 & \textbf{1.65} & 3.36 & 3.45 & 0.82 & 1.41 & 2 & 1.71 & 0 & -- \\
\bottomrule
\end{tabular}
}
\end{table}

Overall, the results show that the classifiers maintain good performance even when applied to data from different segmentation methods. In particular, the Logistic Regression classifier trained on ERGC segments achieved the highest average \textit{cross-method} accuracy ($85.24\%$) and the lowest standard deviation ($1.65$), according to Table \ref{tab:cross_segmenters_extended}, indicating greater stability across tests. It is important to highlight that even after the \textit{cross-method} scenario, considering the accuracy values obtained, a slight improvement can be seen. However, given the average values and standard deviations, it can be seen that the best classifiers tend to present very similar values. 

\textcolor{black}{It is important to emphasize that, although the top-1 classifiers of each superpixel method did not show statistically significant differences in balanced accuracy, the computational cost associated with each segmentation technique varies considerably. In one of the analyzed areas, for instance, CRS required 3.99s to complete the processing, whereas ERGC took 1.7 seconds, ETPS only 0.33 seconds, RSS 0.48 seconds, and SLIC 0.80 seconds. Therefore, ETPS demonstrated the best performance in terms of execution time, being approximately 12 times faster than CRS, which proved to be the most computationally demanding method.}

\textcolor{black}{Furthermore, it was observed that the CRS superpixel method produced segments that were easier to classify, regardless of the training set used, even when it was generated from any of the other methods evaluated in this study. This behavior indicates greater consistency of CRS in supporting the classification stage. In addition, the pair LR/ERGC achieved the highest balanced accuracy average among all the combinations analyzed, showing a slight advantage of this pair compared to the others.}

\begin{figure}[h!]
\centering 
\includegraphics[width=0.8\textwidth]{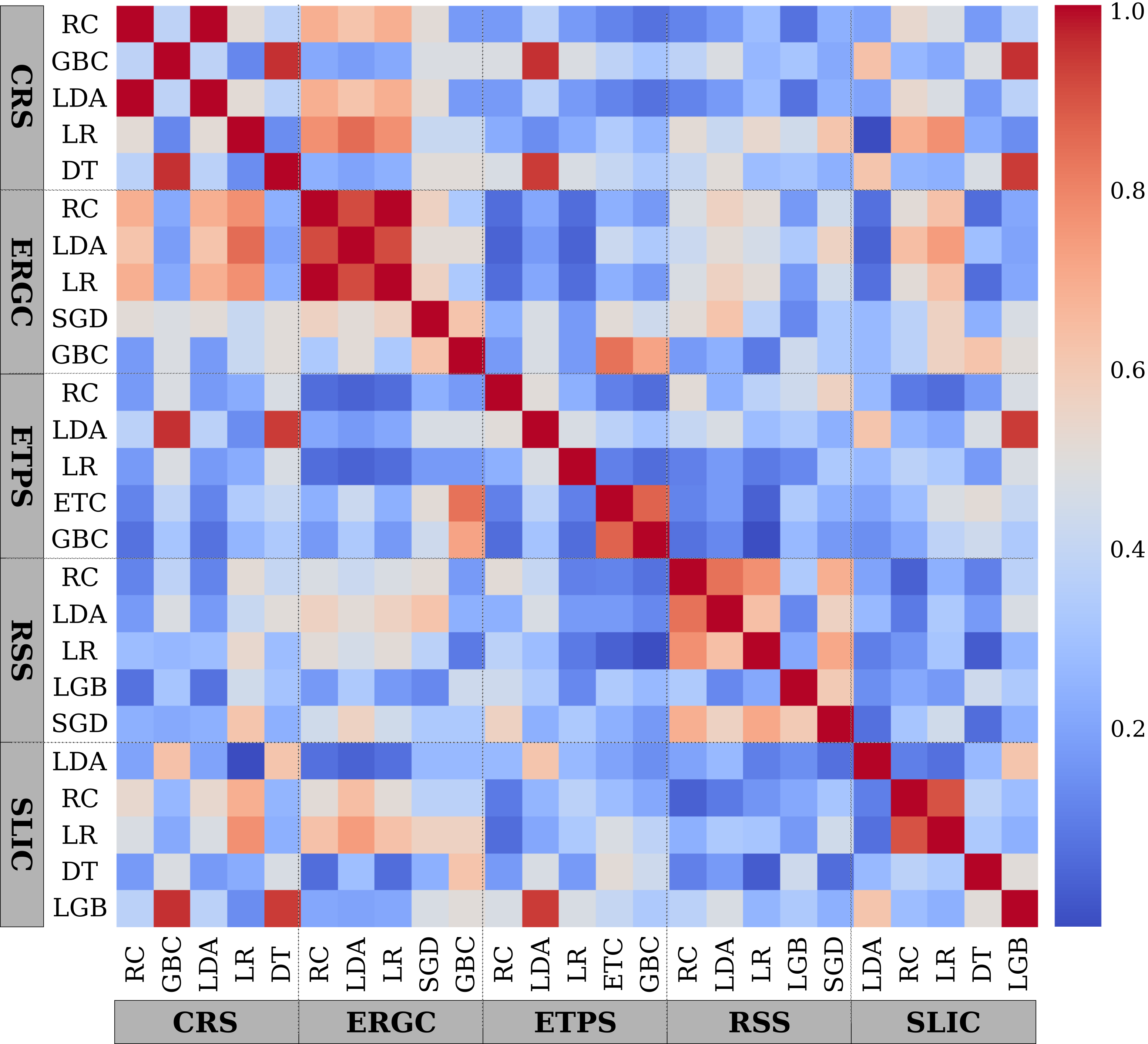}
\caption{Relationship matrix computed among all $25$ classifiers using $COR(c_i,c_j)$ measure.}
\label{fig:corr}
\end{figure}

Since this fact was observed, the next stage of the experiments consisted of verifying the existence of possible complementarity of information between the classifiers composed of different segmentation methods. Therefore, in this work, the diversity measure known as the \textit{correlation coefficient $\rho$} ($COR(c_i,c_j)$) is computed. Given  
{$COR(c_i,c_j) =\frac{ad-bc}{\sqrt{(a+b)(c+d)(a+c)(b+d)}}$, the value $a$ is the percentage of examples that both classifiers $c_i$ and $c_j$ classified correctly in a validation set.
Values $b$ and $c$ are the percentage of examples that $c_j$ hit and
 $c_i$ missed and vice-versa. The value $d$ is the percentage of examples that both classifiers missed.} A lower score between a pair of classifiers $c_i$ and $c_j$ indicates higher diversity~\cite{kuncheva03}. As can be seen in Figure~\ref{fig:corr}, most $COR(c_i,c_j)$ scores are below 0.4 (shades of blue), suggesting that the individual classifiers provide complementary information. Therefore, the use of classifier fusion strategies (ensemble of classifiers) may be a viable way to improve classification results on the target task.

Finally, for the \textbf{third and last stage} of the experiments, a framework for building ensembles of classifiers proposed in the literature has been employed~\cite{Ferreira2021ensembles}. This framework brings together many bioinspired optimization algorithms (e.g., Particle Swarm Optimization, Ant Colony Optimization, Genetic Algorithms, and Artificial Bee Colony algorithms) for classifier selection and fusion. Among the algorithms available in this framework is the Univariate Marginal Distribution Algorithm (UMDA), which achieved the best performance across all comparisons and was selected as the most effective classifier ensemble strategy. UMDA is a type of Estimation of Distribution Algorithm (EDA) that generates new individuals by sampling from univariate normal distributions, which are updated in each iteration based on the top-performing individuals from the previous generation.

Table~\ref{tab:cross_segmenters_extended} also presents the balanced accuracy results obtained by the UMDA strategy, alongside the well-known majority voting (MV) fusion method from the literature. Additionally, the number of classifiers ($\#$C) selected by UMDA is reported. Overall, the UMDA strategy outperformed individual classifiers, achieving a relative gain of up to $2.2\%$. Moreover, it outperformed the MV strategy while using significantly fewer classifiers, less than 50\% of the total used by MV.

\textcolor{black}{A final analysis of the impact of the superpixel methods that compose the ensemble of classifiers based on UMDA strategy can be observed in Table \ref{tab:segmentadores_ensemble}. Among the main findings, it should be noted that the LC/RSS classifier showed a probability of 100\% ($CI$) being selected for the ensemble of classifiers, even when evaluating samples from different superpixel methods. Furthermore, when analyzing the contribution of classifiers per superpixel method ($CS$), it is observed that, in the case of SLIC, its classifiers reached a probability of 60\% in the ensemble of classifiers. Finally, although not explicitly reported in Table \ref{tab:segmentadores_ensemble}, it was verified that all classifiers were selected at least once using the UMDA strategy to compose the final classification solution of some superpixel methods, which reinforces the idea of complementarity among them.}

\begin{table}[ht]
\centering
\caption{Top-3 classifiers/superpixel method with the highest probabilities (Pr., in \%) of being part of the ensemble of classifiers based on UMDA strategy. The probabilities refer to each classifier isolated by superpixel method ($CI$) and to all classifiers of a given superpixel method ($CS$).}
\label{tab:segmentadores_ensemble}
\begin{tabular}{|c|c|c|c|c|c|c|c|c|c|c|c|c|c|c|c|}
\hline
\textbf{Pr.} & \multicolumn{3}{|c|}{\textbf{CRS}} & 
\multicolumn{3}{c|}{\textbf{ERGC}} & 
\multicolumn{3}{c|}{\textbf{ETPS}} & 
\multicolumn{3}{c|}{\textbf{RSS}} & 
\multicolumn{3}{c|}{\textbf{SLIC}} \\
\hline
- & \textbf{GBC} & \textbf{LR} & \textbf{DT} & \textbf{LC} & \textbf{LDA} & \textbf{GBC} & \textbf{LC} & \textbf{LR} & \textbf{ETC} & \textbf{LC} & \textbf{LGB} & \textbf{SGD} & \textbf{LDA} & \textbf{RC} & \textbf{DT} \\
\hline
\textbf{$CI$} & 60 & 60 & 40 & 40 & 40 & 80 & 60 & 40 & 80 & 100 & 40 & 40 & 60 & 80 & 80 \\
\hline
\textbf{$CS$} & \multicolumn{3}{|c|}{40} & 
\multicolumn{3}{c|}{40} & 
\multicolumn{3}{c|}{48} & 
\multicolumn{3}{c|}{44} & 
\multicolumn{3}{c|}{60} \\
\hline
\end{tabular}
\end{table}

\section{Conclusion}

This study investigated the influence of five superpixel-based segmentation methods on training machine learning models for classifying remote sensing images focused on deforestation detection. Based on the top-performing methods identified in previous studies \cite{borlido2024how} and \cite{resende2025igarsstoapper}, the analysis was conducted in three stages: (i) ranking and fine-tuning the top five classifiers for each segmentation method using the PyCaret library; (ii) evaluating generalization through cross-method testing, where classifiers trained on one method were tested on data segmented by others; and (iii) applying ensemble strategies based on bioinspired algorithms to improve accuracy through model combination.

The results showed that, despite known differences among the segmentation methods in terms of perceptual behavior \cite{resende2025igarsstoapper}, all achieved similar performance in supporting model training, demonstrating their suitability for remote sensing tasks. This apparent equivalence raised an important question regarding the need to continue investigating segmentation methods even when they produce similar accuracies.

To address this question, we analyzed the diversity among classifiers using the correlation $COR(c_i,c_j)$ metric. The results (Fig. \ref{fig:corr}) revealed that classifiers based on different segmentation methods capture complementary patterns, which would not be explored if only a single method were used. To demonstrate the benefits of this diversity, we conducted experiments with classifier ensembles, which confirmed that combining segmentation methods improves detection performance in tropical deforestation areas. It is important to note that the ensemble is not a core contribution of this work but serves as evidence that maintaining segmentation diversity is essential to fully exploit the potential of the data.

As future works, we plan a new citizen science campaign using data segmented by the same five methods investigated in this study. The goal is to compare the effectiveness of volunteer annotations across different segmentation methods using PRODES reference labels as ground truth. This will allow a deeper analysis of the impact of segmentation quality on the accuracy of human annotations. \textcolor{black}{Furthermore, it is planned to combine segmentation masks from different segmenters in order to generate segments that better capture complementary spatial patterns and boundaries, thus providing richer representations for subsequent classification tasks. Finally, we highlight that future investigations may include additional classifier performance metrics, such as Precision, Recall, F1-score, AUC-ROC, AUC-PR, MCC, and Cohen’s Kappa, to complement the analysis presented in this study.}

\begin{credits}


\subsubsection{\ackname} The authors would like to thank the UNIFESP and the IFSULDEMINAS for their support; the Coordination for the Improvement of Higher Education Personnel (CAPES) for financial assistance; and the National Laboratory for Scientific Computing (LNCC) for providing the HPC computational resources of the Santos Dumont Supercomputer (SDumont). This work was partially financially supported by the Conselho Nacional de Desenvolvimento Cient{\'i}fico e Tecnol{\'o}gico -- CNPq -- (Grants 306573/2022-9, 442950/2023-3 and 407242/2021-0), the Funda\c{c}{\~a}o de Amparo a Pesquisa do Estado de Minas Gerais -- FAPEMIG -- (APQ-01079-23 and APQ-05058-23). This research is part of INCT Future Internet for Smart Cities, funded by CNPq ($\#$465446/2014-0), and FAPESP ($\#$2014/50937-1, $\#$2015/24485-9, $\#$2017/25908-6, $\#$2018/23908-1, $\#$2019/26702-8, $\#$2023/00811-0,  $\#$2023/00782-0, $\#$2024/01115-0).

\subsubsection{\discintname}
The authors have no competing interests to declare that are relevant to the content of this article.
\end{credits}
%
%
%
\bibliographystyle{splncs04}
\bibliography{refs}

\end{document}